\documentclass[sigconf]{acmart}

\copyrightyear{2024}
\acmYear{2024}
\setcopyright{acmlicensed}\acmConference[CIKM '24]{Proceedings of the 33rd ACM International Conference on Information and Knowledge Management}{October 21--25, 2024}{Boise, ID, USA}
\acmBooktitle{Proceedings of the 33rd ACM International Conference on Information and Knowledge Management (CIKM '24), October 21--25, 2024, Boise, ID, USA}
\acmDOI{10.1145/3627673.3679684}
\acmISBN{979-8-4007-0436-9/24/10}

\usepackage[utf8]{inputenc} 
\usepackage[T1]{fontenc}    
\usepackage{hyperref}    
\usepackage{url}    
\usepackage{booktabs}  
\usepackage{amsfonts} 
\usepackage{nicefrac} 
\usepackage{microtype}
\usepackage{xcolor}
\usepackage{graphicx}
\usepackage{multirow}
\usepackage{threeparttable}
\usepackage{tabularx}
\usepackage{bm}
\usepackage{algorithm}
\usepackage{algorithmic}
\usepackage{enumitem}
\usepackage{subfigure}

\begin{document}

\title{MMPolymer: A Multimodal Multitask Pretraining Framework for Polymer Property Prediction}

\author{Fanmeng Wang}
\authornote{Work done during an internship at DP Technology}
\orcid{0009-0002-2287-2339}
\affiliation{
  \department{Gaoling School of Artificial Intelligence}
  \institution{Renmin University of China}
  \city{Beijing}
  \country{China}
}
\email{fanmengwang@ruc.edu.cn}

\author{Wentao Guo}
\orcid{0000-0001-8058-8323}
\affiliation{
  \department{Department of Chemistry}
  \institution{University of California, Davis}
  \city{Davis}
  \country{United States}
}
\email{wtguo@ucdavis.edu}

\author{Minjie Cheng}
\orcid{0000-0003-1344-5749}
\affiliation{
  \department{Gaoling School of Artificial Intelligence}
  \institution{Renmin University of China}
  \city{Beijing}
  \country{China}
}
\email{chengminjie@ruc.edu.cn}

\author{Shen Yuan}
\orcid{0009-0008-4238-0538}
\affiliation{
  \department{Gaoling School of Artificial Intelligence}
  \institution{Renmin University of China}
  \city{Beijing}
  \country{China}
}
\email{shenyuan721@ruc.edu.cn}

\author{Hongteng Xu}
\authornote{Corresponding author}
\orcid{0000-0003-4192-5360}
\affiliation{
  \department{Gaoling School of Artificial Intelligence}
  \institution{Renmin University of China}
  \city{Beijing}
  \country{China}
}
\email{hongtengxu@ruc.edu.cn}

\author{Zhifeng Gao}
\authornotemark[2]
\orcid{0000-0001-8433-999X}
\affiliation{
  \institution{DP Technology}
  \city{Beijing}
  \country{China}
}
\email{gaozf@dp.tech}

\renewcommand{\shortauthors}{Fanmeng Wang et al.}

\begin{abstract}
Polymers are high-molecular-weight compounds constructed by the covalent bonding of numerous identical or similar monomers so that their 3D structures are complex yet exhibit unignorable regularity. 
Typically, the properties of a polymer, such as plasticity, conductivity, bio-compatibility, and so on, are highly correlated with its 3D structure. 
However, existing polymer property prediction methods heavily rely on the information learned from polymer SMILES sequences (P-SMILES strings) while ignoring crucial 3D structural information, resulting in sub-optimal performance.
In this work, we propose MMPolymer, a novel multimodal multitask pretraining framework incorporating polymer 1D sequential and 3D structural information to encourage downstream polymer property prediction tasks.
Besides, considering the scarcity of polymer 3D data, we further introduce the "Star Substitution" strategy to extract 3D structural information effectively.
During pretraining, in addition to predicting masked tokens and recovering clear 3D coordinates, MMPolymer achieves the cross-modal alignment of latent representations. 
Then we further fine-tune the pretrained MMPolymer for downstream polymer property prediction tasks in the supervised learning paradigm.
Experiments show that MMPolymer achieves state-of-the-art performance in downstream property prediction tasks.
Moreover, given the pretrained MMPolymer, utilizing merely a single modality in the fine-tuning phase can also outperform existing methods, showcasing the exceptional capability of MMPolymer in polymer feature extraction and utilization.
\end{abstract}



\maketitle

\section{Introduction}
In the past few decades, polymers have played a crucial role in many scientific fields, such as chemistry~\cite{carraher2017introduction}, material science~\cite{brazel2012fundamental}, drug design~\cite{pasut2007polymer}, and bioinformatics~\cite{schaffert2008gene}. 
In the applications mentioned above, achieving accurate prediction of polymer properties has garnered more and more attention due to its significance~\cite{audus2017polymer,chen2021polymer}.
For example, predicting polymer properties like plasticity and conductivity helps guide the design and development of polymer-based materials with specific functionalities~\cite{bicerano2002prediction}.
Besides, predicting polymeric drug carriers' bio-compatibility and release kinetics is essential for designing effective and safe drugs~\cite{qiu2006polymer}.

\begin{figure*}[t]
    \centering
    \includegraphics[height=6.2cm]{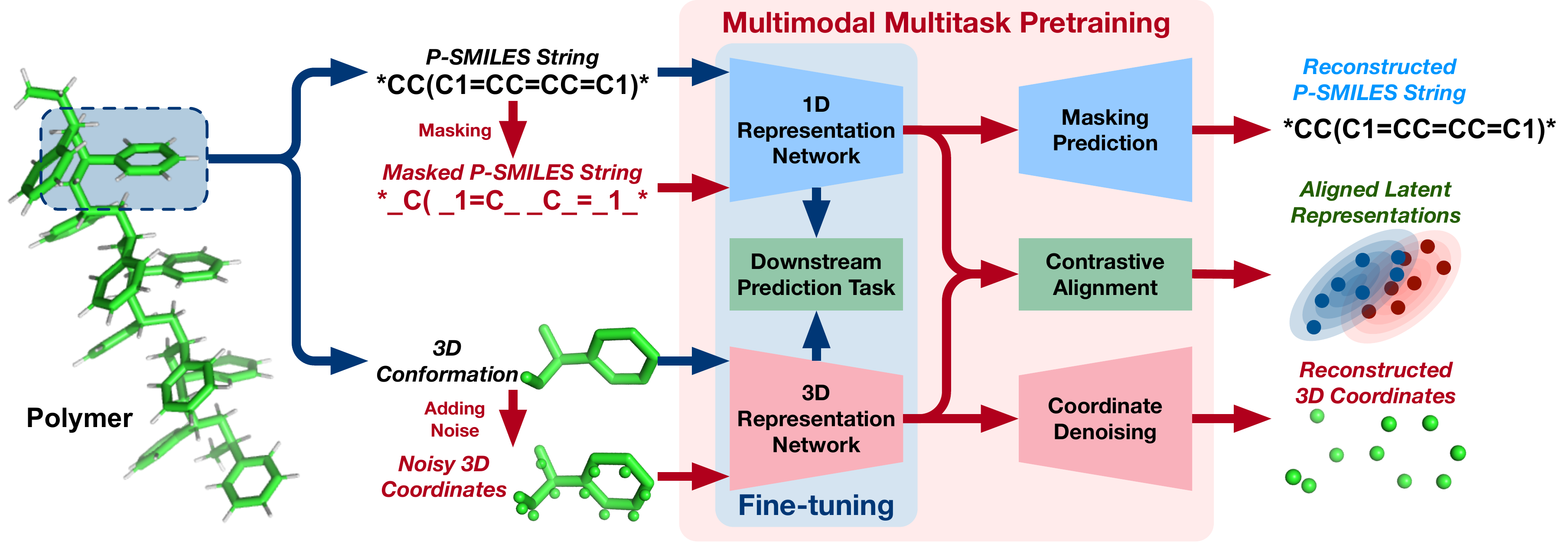}
    \caption{The scheme of the proposed method. Here, the red arrows indicate the pipeline of our multimodal multitask pretraining paradigm, and the blue arrows indicate the pipeline of fine-tuning steps for downstream polymer property prediction tasks. 
    The blue modules are designed for 1D sequences (i.e., P-SMILES strings), and the red modules are designed for 3D conformations. 
    The modules shared by 1D and 3D representations are labeled in green.}
    \label{fig:scheme}
\end{figure*}

Traditional studies on polymers rely on experiments or simulations to assess their properties~\cite{huan2020polymer, danielsen2021molecular}.
While these techniques yield precise outcomes, they suffer from poor scalability due to their high costs, extensive time requirements, and the need for specialized tools and knowledge. 
As a result, they fail to meet the rapidly increasing demands of polymer property prediction.
Subsequently, although some learning-based methods~\cite{roy2006polymer, doan2020machine,amamoto2022data,wang2023scientific} have been proposed for polymer property prediction, their performance is often unsatisfactory due to the scarcity of high-quality polymer property data. 
Recently, some attempts have been made to mitigate data insufficiency based on pretraining techniques~\cite{xu2023transpolymer, kuenneth2023polybert}. 
However, these attempts merely rely on the information from polymer SMILES sequences (i.e., P-SMILES strings) while ignoring the fact that the properties of a polymer are highly correlated with its 3D structure, which leads to sub-optimal performance as well.

To overcome the limitations of existing polymer property prediction methods, we propose a novel multimodal multitask pretraining framework, MMPolymer, leveraging 3D structural information to enhance the predictive capabilities of polymer properties. 
Here, considering the regularity of polymer 3D structure and scarcity of polymer 3D data, we leverage the 3D conformation of the corresponding repeating unit\footnote{Not equivalent to the monomer, which lacks polymerization site information.} to approximate the whole polymer 3D conformation.
Specifically, we first convert the corresponding P-SMILES string through our "Star Substitution" strategy, where the "*" symbol in the P-SMILES string is replaced by the neighboring atom symbol of another "*" symbol. Then we use the RDKit~\cite{landrum2013rdkit} tool to generate the 3D conformation based on the converted P-SMILES string.
In this way, the generated 3D conformation can not only reflect 3D structural features within repeating units but also reflect 3D structural features between the repeating units, thus contributing to extracting polymer 3D structural information effectively.

During pretraining, we mask the P-SMILES string randomly, add noise to the atom coordinates of corresponding 3D conformation, and pass them through the 1D and 3D representation networks, respectively. 
As illustrated in Figure~\ref{fig:scheme}, we learn the proposed representation networks based on three pretraining tasks, including predicting masked tokens, recovering clear 3D coordinates (i.e., coordinate denoising), and aligning the latent representations across the two modalities by contrastive learning~\cite{liu2021self}.
Solving these three tasks jointly leads to the proposed multimodal multitask pretraining paradigm, which helps our model fully incorporate polymer 1D sequential and 3D structural information. 
Then we further fine-tune the pretrained MMPolymer in the supervised learning paradigm for downstream polymer property prediction tasks.

\textbf{To the best of our knowledge, MMPolymer is the first work that incorporates 3D structural information into polymer property prediction.}
The experimental results on various polymer property datasets consistently demonstrate that MMPolymer significantly outperforms existing polymer property prediction methods, achieving state-of-the-art performance. 
Moreover, even merely relying on single-modal information (either polymer 1D sequential or 3D structural information) during fine-tuning, MMPolymer still outperforms existing polymer property prediction methods, showcasing its promising capability of extracting and utilizing polymer features.
Besides, comprehensive ablation studies are also conducted to offer valuable insights into the performance of MMPolymer, further supporting its rationality and effectiveness.

\begin{figure*}[t]
    \centering 
    \includegraphics[height=4.5cm]{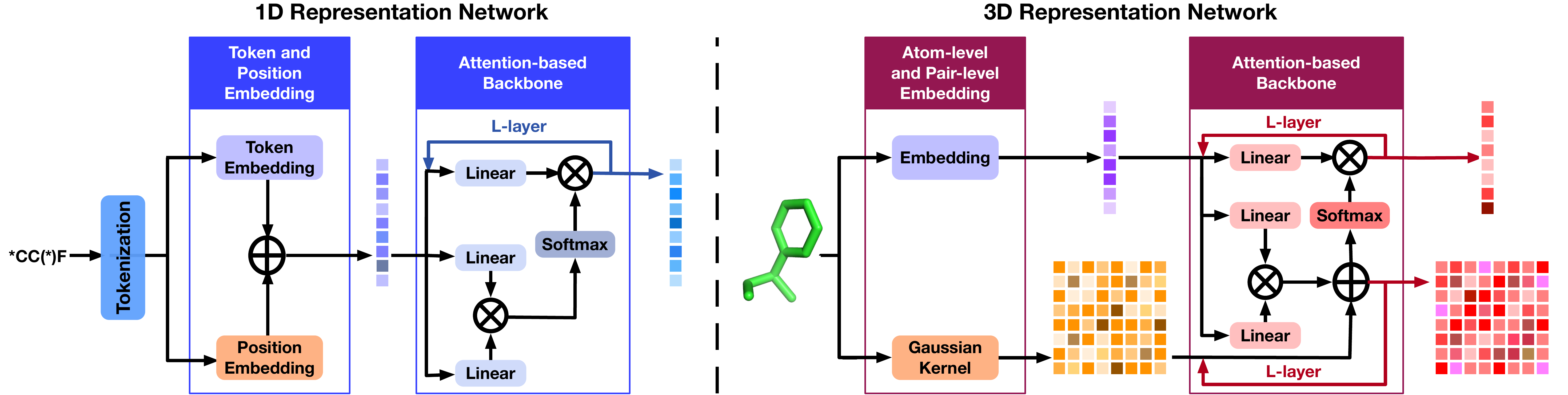} 
    \caption{Left: the architecture of our 1D representation network, which takes polymer SMILES sequences (i.e., P-SMILES strings) as input, and outputs corresponding 1D sequential representation. Right: the architecture of our 3D representation network, which takes 3D conformation as input, and outputs corresponding 3D structural representation.}
    \label{fig:model_archi}  
\end{figure*}

\section{Related Work}
\subsection{Polymer Property Prediction}

Polymer property prediction is one of the fundamental problems in polymer informatics, and various machine learning and deep learning algorithms have been widely used as promising tools~\cite{hatakeyama2023recent}.

Early works~\cite{bicerano2002prediction, yu2006prediction} in polymer property prediction employed classic machine learning algorithms (e.g., multiple linear regression) to predict corresponding polymer properties from polymer sequences.
The work in~\cite{le2012quantitative} further applied kernel support vector machines (kernel SVMs) to deal with non-linear relationships between polymer sequences and their properties, achieving great predictive performance.
For deep learning examples, the work in~\cite{rahman2021machine} predicted mechanical properties of polymer-carbon nanotube surfaces using convolutional neural networks (CNNs) while recurrent neural networks
(RNNs) were also used in~\cite{simine2020predicting, webb2020targeted} to learn the latent relationships between polymer sequences and their properties.
However, these methods are significantly limited by the scarcity of high-quality polymer property data.
Recently, inspired by the exceptional performance of various pretrained models in natural language processing (NLP) tasks~\cite{min2023recent, liu2023pre, zhao2023survey}, some Transformer-based pretraining methods have been proposed for polymer property prediction.
These methods like Transpolymer~\cite{xu2023transpolymer} and polyBERT~\cite{kuenneth2023polybert} treat polymers as character sequences (i.e., P-SMILES strings) and undergo pretraining on extensive unlabeled polymer sequences, so they heavily rely on the information learned from polymer sequences.
However, the latest studies ~\cite{stark20223d, zhou2023unimol, wang2023automated} have revealed that properties are mainly determined by the 3D structure, thus highlighting the crucial need for a paradigm shift towards integrating 3D structural information into polymer property prediction.

\subsection{Pretraining for Molecular Modeling}

A closely related domain to polymer science is molecular science~\cite{ma2022disentangled, feng2022mgmae, sun2022pemp}, where numerous pretraining methods have been developed for molecular modeling in the past few years.

SMILES-BERT~\cite{wang2019smiles} and ChemBERTa~\cite{ahmad2022chemberta} were first proposed and achieved great predictive performance on various molecular property prediction tasks by undergoing pretraining on extensive molecular sequences (e.g., SMILES strings).
Then, due to the rapid development of graph neural networks~\cite{satorras2021n, yuan2021semi, wu2022graph},
subsequent works were extended to molecular graphs.
For example, MolCLR~\cite{wang2022molecular} was pretrained on 10 million molecular graphs by contrastive learning strategy, and the work in~\cite{benjamin2022graph} pretrained graph neural networks by utilizing inherent properties of molecular graphs.
Recently, considering that intramolecular interactions are fundamentally three-dimensional~\cite{fang2022geometry, wang2023mperformer, zhu2023learning}, many works tend to incorporate 3D structural information into pretraining to acquire more comprehensive molecular representation.
For example, the work in~\cite{zhu2022unified} proposed a unified 2D and 3D pretraining framework for more informative representation, and 3D Infomax~\cite{stark20223d} proposed to maximize the mutual information between 3D and 2D representation during pretraining to improve the performance of molecular property prediction.

\section{Method}
\subsection{Overview}
Given a polymer, we denote it as a tuple $\{\bm{S},\bm{C}\}$. Here, $\bm{S}$ represents its SMILES sequence (i.e., P-SMILES string), and $\bm{C}=(\bm{A},\bm{P})$ represents the 3D conformation, where $\bm{A}=[a_i]\in\mathbb{N}^N$ is atom types, $\bm{P}=[\bm{p}_i]\in\mathbb{R}^{N\times 3}$ is atom 3D coordinates, and $N$ is the number of atoms.

To acquire a comprehensive polymer representation for downstream polymer property prediction tasks, we pretrain MMPolymer through the multimodal multitask paradigm.
As shown in Figure~\ref{fig:scheme}, 
the 1D representation network $f^{1d}:\mathcal{S}\mapsto \mathcal{X}$ is pretrained via the masked prediction task. 
Meanwhile, to effectively incorporate 3D structural information, the 3D representation network $f^{3d}:\mathcal{C}\mapsto \mathcal{X}$ is pretrained via the coordinate denoising task.
Besides, we further align the latent representations across the two modalities via the cross-modal alignment task during pretraining.
Here, $\mathcal{S}$ is the collection of P-SMILES strings, $\mathcal{C}$ is the collection of 3D conformations, and $\mathcal{X}\subset \mathbb{R}^{d}$ is the latent space.

\subsection{Model Architecture}

\subsubsection{1D Representation Network}\label{section:1D_net}
The 1D representation network $f^{1d}$, based on Transformer architecture~\cite{vaswani2017attention}, takes polymer sequences as input, and outputs the corresponding 1D sequential representation $\bm X_{1d}$, aiming to encode as much information as possible about polymer sequences.

Unlike small molecules, which can be easily represented by corresponding sequences (e.g., SMILES string), converting polymers to sequences is not straightforward due to their complex structures and compositions.
Here, we use the P-SMILES string~\cite{kim2018polymer}, a modified SMILES string for polymers, as input polymer sequences. 
\textbf{Specifically, the P-SMILES string is formed by combining the SMILES string~\cite{o2012towards} of the corresponding monomer with two "*" symbols which are used to indicate the polymerization sites (i.e., the connected atoms between repeating units).}
For example, polyethylene is represented by "*CC*" and polypropylene is represented by "*CC(*)C".
Besides, if other descriptors (e.g., polymerization degree) are also available, they can be added to the end of the corresponding P-SMILES string and fed into our 1D representation network together.

\begin{figure}[t]
    \centering 
    \includegraphics[height=4.1cm]{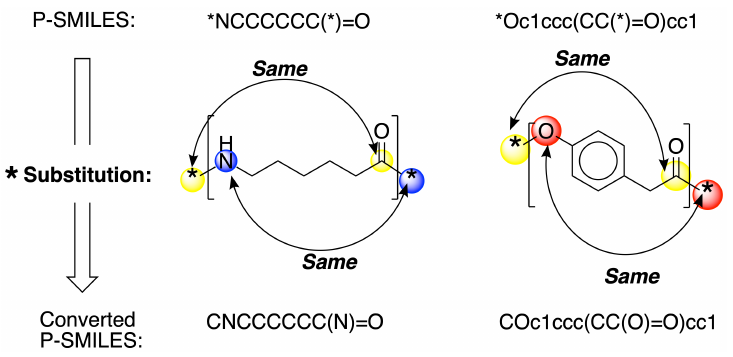} 
    \caption{Visualization of our "Star Substitution" strategy.}
    \label{fig:star_sub}  
\end{figure}

As shown in Figure~\ref{fig:model_archi}, given a polymer sequence represented by the corresponding P-SMILES string "*CC(*)F", we first use chemical-aware tokenization from ~\cite{xu2023transpolymer} to get corresponding tokens.
Here, the P-SMILES string "*CC(*)F" is converted into [`*', `C', `C', `(', `*', `)', `F'] along with special tokens (i.e., [CLS] and [SEP]).
After tokenization, these tokens are further converted into continuous vector representations by adding
each token's embedding to its corresponding position embedding. 
As a result,  these vectorized tokens can be effectively interpreted and handled by the attention-based backbone of our 1D representation network.
Specifically, segment embeddings are removed and we use the typical trigonometric function to acquire corresponding position embedding, i.e.,
\begin{eqnarray}
\text{Position\_Embedding}(pos, 2i) = \sin\Bigl(\frac{pos}{10000^{2i/d}}\Bigr),
\end{eqnarray}
\begin{eqnarray}
\text{Position\_Embedding}(pos, 2i+1) = \cos\Bigl(\frac{pos}{10000^{2i/d}}\Bigr),
\end{eqnarray}
where $pos$ is the position in the token sequence, $d$ is the total embedding dimension, and $i$ is the index of the embedding dimension, ranging from $0$ to 
$\frac{1}{d} - 1$.

Finally, the embedding vectors are fed into the attention-based backbone (the same as the one used in Roberta~\cite{liu2019roberta}) of our 1D representation network to get the corresponding 1D sequential representation $\bm{X}_{1d}$.
Here we utilize the output [CLS] representation as the corresponding 1D sequential representation $\bm X_{1d}$ for the input polymer sequence, i.e.,
\begin{eqnarray}
    \bm{X}_{1d} = Rep_{[CLS]}.
\end{eqnarray}

\subsubsection{3D Representation Network}\label{section:3D_net}
The 3D representation network $f^{3d}$, based on SE(3)-Transformer architecture~\cite{fuchs2020se}, takes 3D conformation as input and outputs corresponding SE(3)-invariant representation vector as 3D structural representation $\bm X_{3d}$, aiming to encode comprehensive polymer 3D structural information.

\textbf{"Star Substitution" strategy.} Due to the regularity of polymer 3D structure and scarcity of polymer 3D data, we approximate the whole polymer 3D conformation with the 3D conformation of its corresponding repeating unit.
Specifically, as shown in Figure~\ref{fig:star_sub}, we first generated the converted P-SMILES string through our "Star Substitution" strategy, where \textbf{the "*" symbol in the P-SMILES string is replaced by the neighboring atom symbol of another "*" symbol.}
For example, "*NCCCCCC(*)=O" is converted to "CNCCCCCC(N)=O" and "*Oc1ccc(CC(*)=O)cc1" is converted to "COc1ccc(CC(O)=O)cc1".
Then we further generate the corresponding 3D conformation by RDKit~\cite{landrum2013rdkit} based on the converted P-SMILES string. 
In this way, the generated 3D conformation can not only reflect 3D structural features within repeating units but also reflect 3D structural features between the repeating units, thus contributing to approximating the whole polymer 3D conformation as much as possible.

As illustrated in Figure~\ref{fig:model_archi}, the 3D conformation $\bm{C} = (\bm{A}, \bm{P})$ of the corresponding repeating unit is used as input of our 3D representation network, where $\bm{A}=[a_i]\in\mathbb{N}^N$ is atom types, $\bm{P}=[\bm{p}_i]\in\mathbb{R}^{N\times 3}$ is atom 3D coordinates and $N$ is the number of atoms.
Besides, the atom $a_0$ is a virtual atom, analogous to the role of [CLS] in natural language processing tasks, and its 3D coordinate $\bm p_0$ is located at the space center of the whole 3D conformation.

Here, we first use an embedding layer to encode atom types $\bm{A}$ to the initial atom-level representation $\bm{X}^{(0)}_{a} \in \mathbb{R}^{N \times d_a}$ as follows:
\begin{eqnarray}\label{eq:atom_rep}
\bm{X}^{(0)}_{a} = f_a (\bm{A}),
\end{eqnarray}
where $f_a$ embeds each atom type to a $d_a$-dimensional embedding.

Meanwhile, a pair-type aware Gaussian kernel~\cite{scholkopf1997comparing} encodes the Euclidean distances of each atom pair $(a_i, a_j)$ as the initial pair-level representation $\bm{X}^{(0)}_{p} \in \mathbb{R}^{N \times N \times d_p}$, i.e.,
\begin{eqnarray}\label{eq:pair_rep}
\bm{X}^{(0)}_{p}[i,j] = \mathcal{G}(\bm v_{{a_i}{a_j}} \|\bm{p}_i - \bm{p}_j\| + \bm u_{{a_i}{a_j}} - \bm \mu,  \bm \sigma).
\end{eqnarray}
where $\bm v_{{a_i}{a_j}} \in \mathbb{R}^{d_p}$ and $\bm u_{{a_i}{a_j}} \in \mathbb{R}^{d_p}$ represent the weights and biases, which are learnable parameters for each atom pair $(a_i, a_j)$, 
$\mathcal{G}$ is a Gaussian basis kernel function with mean value $\bm \mu \in \mathbb{R}^{d_p}$ and standard deviation $\bm \sigma \in \mathbb{R}^{d_p}$.

Then, the initial representation $\bm{X}^{(0)}_{a}$ and $\bm{X}^{(0)}_{p}$ are fed into the attention-based backbone (the same as the one used in Uni-Mol~\cite{zhou2023unimol}) of our 3D representation network to get the final representation $\bm{X}^{(L)}_{a}$ and $\bm{X}^{(L)}_{p}$.
To fully capture intricate 3D structure information, the multi-head self-attention mechanism is applied to the encoder layers, and the number of attention heads is determined by the dimension of pair-level embedding $d_p$.
Besides, to update atom-level and pair-level representation effectively, we use atom-to-pair communication to improve the traditional self-attention mechanism.
The specific implementation of the $h$-th attention head in the $l$-th encoder layer can be expressed as follows:
\begin{eqnarray}
    \bm{Atten}^{(l,h)} = \frac{\bm{Q}^{(l,h)} (\bm{K}^{(l,h)})^{\top} }{\sqrt {d_h}} + \bm{X}^{(l-1,h)}_{p},
\end{eqnarray}
\begin{eqnarray}
    \bm{X}^{(l,h)}_{a} = \sigma (\bm{Atten}^{(l,h)}) \bm{V}^{(l,h)}, \quad \bm{X}^{(l,h)}_{p} = \bm{Atten}^{(l,h)},
\end{eqnarray}
where $\bm{Q}^{(l,h)} = \bm{X}^{(l-1)}_{a} \bm{W}_Q^{(l,h)}$, $\bm{K}^{(l,h)} = \bm{X}^{(l-1)}_{a} \bm{W}_K^{(l,h)}$, and $\bm{V}^{(l,h)} = \bm{X}^{(l-1)}_{a} \bm{W}_V^{(l,h)}$ are query, key, and value matrices generated by corresponding linear maps $\bm{W}_Q^{(l,h)},\bm{W}_K^{(l,h)},$ $ \bm{W}_V^{(l,h)}\in \mathbb{R}^{d_a \times d_h}$.
Beisdes, $\bm{X}^{(l-1,h)}_{p}\in \mathbb{R}^{N \times N}$ is the $h$-th slice of  $\bm{X}^{(l-1)}_{p} \in \mathbb{R}^{N \times N \times d_p}$, $\sigma(\cdot)$ is the row-wise softmax operator, and $d_h=d_a/d_p$ is the hidden dimension of each attention head.

By concatenating the outputs of each attention head, we can further get the outputs of the whole $l$-th layer as follows:
\begin{eqnarray}
    &\bm{X}^{(l)}_{a} = \text{Concat}(\{\bm{X}^{(l,h)}_{a}\}_{h=1}^{d_p})\bm{W}_{O}^{(l)} + \bm{X}^{(l-1)}_{a},
\end{eqnarray}
\begin{eqnarray}
    \bm{X}^{(l)}_{p} = \text{Concat}(\{\bm{X}^{(l,h)}_{p}\}_{h=1}^{d_p}),
\end{eqnarray}
where $\text{Concat}(\cdot)$ represents the concatenation operator, and $\bm{W}_O^{(l)} \in \mathbb{R}^{d_a \times d_a}$ represents the feed-forward neural network.

After $L$ encoder layers, we finally get the 3D structural representation $\bm{X}_{3d}$ from the final atom-level representation of the corresponding virtual atom $a_0$ as follows:
\begin{eqnarray}
    \bm{X}_{3d} = \bm{X}^{(L)}_{a}[0,:].
\end{eqnarray}

\subsection{Learning Paradigm}\label{section:learning_paradigm}
As shown in Figure~\ref{fig:scheme}, we pretrain our MMPolymer with a multimodal multitask paradigm, including the masked prediction task for pretraining the 1D representation network $f^{1d}$ based on polymer 1D sequence data in Sec.~\ref{section:MLM_task}, the coordinate denoising task for pretraining the 3D representation network $f^{3d}$ based on polymer 3D structure data in Sec.~\ref{section:3D_task}, and the cross-modal alignment task for aligning the learned latent representations across the two modalities in Sec.~\ref{section:contras_task}, to acquire a comprehensive polymer representation for downstream polymer property prediction tasks.
The total loss during pretraining can be expressed as follows:
\begin{equation}
\mathcal{L}_{\text{pretrain}} = \mathcal{L}_{\text{1D}} + \mathcal{L}_{\text{3D}} + \mathcal{L}_{\text{contrastive}}. 
\end{equation}

After pretraining, we further fine-tune the pretrained MMPolymer on various polymer property datasets for the corresponding polymer property
prediction tasks in the supervised learning paradigm, as described in Sec.~\ref{section:finetune}.

\subsubsection{Masked Prediction Task}\label{section:MLM_task}
During the masked prediction task (i.e., masked language modeling task in NLP), approximately 15\% of tokens in a polymer sequence are randomly selected for masking. 
Subsequently, these chosen tokens are subjected to three possible replacement options: they may be replaced with a special token [MASK], a random token, or left unchanged. 
Then, the 1D representation network $f^{1d}$ is trained to recover the original identity of these masked tokens based on the contextual information provided by the surrounding sequence. This task encourages our 1D representation network $f^{1d}$ to capture the nuanced structures and interactions in polymer sequences.

Specifically, $\mathcal{V}$ is the vocabulary set, $\mathcal{M}$ is the set of masked positions, $\bm {\hat Y} = [\bm{\hat{y}}_i] \in \mathbb{R}^{|\mathcal{M}| \times |\mathcal{V}|}$ is the predicted probability distribution over these masked positions, and $\bm {Y} = [\bm{{y}}_i] \in \mathbb{R}^{|\mathcal{M}| \times |\mathcal{V}|}$ represents the corresponding label in the one-hot format.
Here, we utilize the cross-entropy loss~\cite{zhang2018generalized} as our loss function for the masked prediction task, i.e.,
\begin{equation}
\mathcal{L}_{\text{1D}} = -\frac{1}{|\mathcal{M}|} \sideset{}{_{i \in \mathcal{M}}}\sum \sideset{}{_{j}^{|\mathcal{V}|}}\sum \bm{y_i}[j] \cdot \log(\bm {\hat y_i}[j]),
\end{equation}

Through the above process, our 1D representation network $f^{1d}$ acquires valuable insights into the complex patterns and relationships within polymer sequences, thereby greatly benefiting downstream polymer property prediction tasks.

\subsubsection{Coordinate Denoising Task}\label{section:3D_task}

During the coordinate denoising task, we first randomly add noise to the atom coordinates of the given 3D conformation $\bm C$, which are combined by atom types $\bm{A} \in \mathbb{N}^{N}$ and atom 3D coordinates $\bm{P} \in \mathbb{R}^{N \times 3}$. 
This process generates corrupted 3D conformation $\bm{\tilde{C}}$ with noisy 3D atom coordinates $\bm{\tilde {P}}$, which serves as input of our 3D representation network $f^{3d}$.
Then the 3D representation network $f^{3d}$ is trained to recover the original 3D atom coordinates $\bm P$ from the corrupted 3D conformation $\bm{\tilde{C}}$.

As described Sec.~\ref{section:3D_net}, the corrupted 3D conformation $\bm{\tilde{C}}$ is first converted to initial atom-level representation $\tilde{ \bm{X}}^{(0)}_{a} \in \mathbb{R}^{N \times d_a}$ and pair-level representation $\tilde {\bm{X}}^{(0)}_{p} \in \mathbb{R}^{N \times N \times d_p}$. 
Then $\tilde{ \bm{X}}^{(0)}_{a}$  and $\tilde{ \bm{X}}^{(0)}_{p}$ are fed into our attention-based backbone to get the final atom-level representation $\tilde{ \bm{X}}^{(L)}_{a}$ and pair-level representation $\tilde {\bm{X}}^{(L)}_{p}$ through $L$ encoder layer.
Finally, we use a pair-level decoder to recover the original 3D atom coordinates based on the learned pair-level representation.
Here, we utilize the smooth L1 loss~\cite{wang2020comprehensive} as our loss function for the coordinate denoising task, i.e.,
\begin{eqnarray}\label{eq:rec}
\bm{\hat{p}}_i = \tilde{\bm{p}}_i + \sideset{}{_{j=1}^{N}}\sum \frac{\psi(\tilde{\bm{x}}^{(L)}_{p,ij} - \tilde{\bm{x}}^{(0)}_{p,ij})(\tilde{\bm{p}}_i - \tilde{\bm{p}}_j)}{N}, 
\end{eqnarray}
\begin{eqnarray*}\label{eq:rec_loss}
\mathcal{L}_{\text{3D}} = \frac{1}{N} \sum_{i=1}^{N}\sum_{j=1}^{3}
\begin{cases}
0.5 | {\bm p}_{i}[j] - \bm{\hat p}_{i}[j] |^2, & \text{if } | {\bm p}_{i}[j] - \bm{\hat p}_{i}[j] | < 1, \\
| {\bm p}_{i}[j] - \bm{\hat p}_{i}[j] | - 0.5, & \text{otherwise}.
\end{cases}
\end{eqnarray*}
where $N$ is the number of atoms, $\tilde{\bm{p}}_i \in\mathbb{R}^{3}$ is the noisy 3D coordinate of $i$-th atom, $\hat{\bm{p}}_i \in\mathbb{R}^{3}$ is the predicted 3D coordinate of $i$-th atom, $\psi$ is an MLP layer that maps the update of pair-level representation to a scalar as weight, while
$\tilde{\bm{x}}^{(L)}_{p,ij} \in\mathbb{R}^{d_p}$ and $\tilde{\bm{x}}^{(0)}_{p,ij}\in\mathbb{R}^{d_p}$ are final and initial pair-level representations of atom pair $(i,j)$.

Generally, the coordinate denoising task enables our 3D representation network $f^{3d}$ to accurately capture essential polymer 3D structural information, which is crucial for downstream polymer property prediction tasks.

\subsubsection{Cross-modal Alignment Task}\label{section:contras_task}
During pretraining, we also employ contrastive learning to align 1D sequential representation $\bm X_{1d}$ learned by 1D representation network $f^{1d}$ and 3D structural representation $\bm X_{3d}$ learned by 3D representation network $f^{3d}$.

Specifically, we derive the corresponding multimodal representations $\{\bm X_{1d}^{i}, \bm X_{3d}^{i}\}_{i=1}^{K}$ for a batch of $K$ polymers. 
Our goal is to maximize the similarity between positive multimodal representation pairs from the same polymer and minimize the similarity between negative multimodal representation pairs from different polymers. 
Here, we utilize the classical InfoNCE loss~\cite{oord2018representation}, as our loss function for the cross-modal alignment task, i.e.,
\begin{equation}
\mathcal{L}_{\text{contrastive}} = -\frac{1}{K} \sideset{}{_{i=1}^{K}}\sum \log\Biggl(\frac{\exp(\text{sim}(\bm X_{1d}^{i}, \bm X_{3d}^{i}) / \tau)}{\sum_{j=1}^{K}\exp(\text{sim}(\bm X_{1d}^{i}, \bm X_{3d}^{j}) / \tau)}\Biggr),
\end{equation}
where $K$ is the batch size, $\text{sim}(\cdot)$ is the cosine similarity function, and $\tau$ is the temperature parameter.

Through the contrastive learning task during pretraining, the 1D sequential representation and 3D structural representation of polymers are aligned into a shared space,
enhancing the coherence and mutual informativeness of these multimodal representations.

\subsubsection{Fine-tuning}\label{section:finetune}
Benefiting from our multimodal multitask pretraining paradigm, MMPolymer can learn comprehensive polymer representation after pretraining.
Subsequently, we further fine-tune the pretrained MMPolymer through a prediction head (i.e., a collection of multi-layer perceptions), to achieve accurate polymer property prediction in the supervised learning paradigm.

Moreover, although the pretraining phase is multi-modal, the fine-tuning phase allows for flexible modality information choice by using either a single modality representation (1D sequential representation $\bm X_{1d}$ or 3D structural representation $\bm X_{3d}$) or combining both modalities to predict corresponding polymer properties. 
This choice depends on the characteristics of the property being predicted and the specific requirements of the polymer property prediction task at hand. 
Regardless of the chosen modality representation, the multimodal multitask pretraining paradigm has already equipped our MMPolymer with adequate general knowledge for achieving accurate polymer property prediction.

\section{Experiments}
In this section, to demonstrate the effectiveness of our proposed MMPolymer, we compare it with several state-of-the-art prediction methods on various polymer property datasets.
Besides, comprehensive ablation studies are also conducted to offer valuable insights into the performance of our proposed method.

\begin{table}[tbp]
  \centering
  \caption{The summary of our datasets, where the pretraining dataset is unlabeled and the fine-tuning datasets are regression-type datasets with corresponding property labels.}
    \begin{tabular}{c|c|c|c}
    \toprule
    Dataset & Property & Data Range & Data Size \\
    \midrule
    PI1M  & / & / & $\sim$1M \\
    Egc   & bandgap (chain) & $[0.02, 8.30]$ & 3380 \\
    Egb   & bandgap (bulk) & $[0.39, 10.05]$ & 561 \\
    Eea   & electron affinity & $[0.39, 4.61]$ & 368 \\
    Ei    & ionization energy & $[3.55, 9.61]$ & 370 \\
    Xc    & crystallization tendency & $[0.13, 98.41]$ & 432 \\
    EPS   & dielectric constant & $[2.61, 8.52]$ & 382 \\
    Nc    & refractive index & $[1.48, 2.58]$ & 382 \\
    Eat   & atomization energy & $[-6.83, -5.02]$ & 390 \\
    \bottomrule
    \end{tabular}%
  \label{tab: data}%
\end{table}%

\subsection{Experimental Setup}

\subsubsection{Datasets}
We use the PI1M dataset~\cite{ma2020pi1m}, which contains about one million unlabeled polymer data, to pretrain our MMPolymer.
Then, we employ eight open-source polymer property datasets (denoted as Egc, Egb, Eea, Ei, Xc, EPS, Nc, and Eat, respectively) provided in~\cite{zhang2023transferring} as our fine-tuning datasets\footnote{Although some other polymer property datasets have been mentioned in previous works~\cite{kuenneth2021polymer, kuenneth2023polybert}, these datasets like PolyInfo database~\cite{otsuka2011polyinfo}, are not publicly available, thus making them inaccessible for our experiments.}.
These property datasets, obtained through density functional theory (DFT) calculations~\cite{orio2009density}, encompass a broad range of typical polymer properties and have been extensively utilized in previous works~\cite{zhang2023transferring, xu2023transpolymer}.
More details about our datasets are presented in Table~\ref{tab: data}.

\begin{table*}[h]
\caption{The performance comparison of different methods on eight polymer property datasets, and the best result for each polymer property dataset has been bolded.} 
  \centering
  \setlength{\tabcolsep}{3pt}
    \centering\begin{tabular}{c|c|cccccccc}
    \toprule
    Metric & Method & Egc   & Egb   & Eea   & Ei    & Xc    & EPS   & Nc    & Eat \\
    \midrule
    \multirow{9}[0]{*}{RMSE ($\downarrow$)} 
        & ChemBERTa~\cite{ahmad2022chemberta} & 0.539$_{\pm\text{0.049}}$ & 0.664$_{\pm\text{0.079}}$  & 0.350$_{\pm\text{0.036}}$ & 0.485$_{\pm\text{0.086}}$ & 18.711$_{\pm\text{1.396}}$ & 0.603$_{\pm\text{0.083}}$ & 0.140$_{\pm\text{0.010}}$ & 0.219$_{\pm\text{0.056}}$  \\

        & MolCLR~\cite{wang2022molecular} & 0.587$_{\pm\text{0.024}}$ & 0.644$_{\pm\text{0.072}}$ & 0.404$_{\pm\text{0.017}}$ & 0.533$_{\pm\text{0.053}}$ & 21.719$_{\pm\text{1.631}}$ & 0.631$_{\pm\text{0.045}}$ &
        0.117$_{\pm\text{0.015}}$ & 0.094$_{\pm\text{0.033}}$ \\

        & 3D Infomax~\cite{stark20223d} & 0.494$_{\pm\text{0.039}}$ & 0.553$_{\pm\text{0.032}}$ & 0.335$_{\pm\text{0.055}}$ & 0.449$_{\pm\text{0.086}}$ & 19.483$_{\pm\text{2.491}}$ & 0.582$_{\pm\text{0.054}}$ &
        0.101$_{\pm\text{0.018}}$ & 0.094$_{\pm\text{0.039}}$ \\

        & Uni-Mol~\cite{zhou2023unimol} & 0.489$_{\pm\text{0.028}}$ & 0.531$_{\pm\text{0.055}}$  & 0.332$_{\pm\text{0.027}}$ & 0.407$_{\pm\text{0.080}}$ & {17.414}$_{\pm\text{1.581}}$ & {0.536}$_{\pm\text{0.053}}$ & 0.095$_{\pm\text{0.013}}$ & 0.084$_{\pm\text{0.034}}$  \\
        
        & SML~\cite{zhang2023transferring} & 0.489$_{\pm\text{0.056}}$ & 0.547$_{\pm\text{0.110}}$ & {0.313}$_{\pm\text{0.016}}$ & 0.432$_{\pm\text{0.060}}$ & 18.981$_{\pm\text{1.258}}$ & 0.576$_{\pm\text{0.020}}$ & 0.102$_{\pm\text{0.010}}$ & 0.062$_{\pm\text{0.014}}$ \\
    
        & PLM~\cite{zhang2023transferring} & 0.459$_{\pm\text{0.036}}$ & {0.528}$_{\pm\text{0.081}}$ & 0.322$_{\pm\text{0.037}}$ & 0.444$_{\pm\text{0.062}}$ & 19.181$_{\pm\text{1.308}}$ & 0.576$_{\pm\text{0.060}}$ & 0.100$_{\pm\text{0.010}}$ &{\textbf{0.050}$_{\pm\text{0.010}}$} \\
        
        & polyBERT~\cite{kuenneth2023polybert} & 0.553$_{\pm\text{0.011}}$ & 0.759$_{\pm\text{0.042}}$ & 0.363$_{\pm\text{0.037}}$ & 0.526$_{\pm\text{0.068}}$ & 18.437$_{\pm\text{0.560}}$ & 0.618$_{\pm\text{0.049}}$ & 0.113$_{\pm\text{0.003}}$ & 0.172$_{\pm\text{0.016}}$ \\

        & Transpolymer~\cite{xu2023transpolymer} & {0.453}$_{\pm\text{0.007}}$ & 0.576$_{\pm\text{0.021}}$ & 0.326$_{\pm\text{0.040}}$ & {0.397}$_{\pm\text{0.061}}$ & {17.740}$_{\pm\text{0.732}}$ & {0.547}$_{\pm\text{0.051}}$ & {0.096}$_{\pm\text{0.016}}$ & 0.147$_{\pm\text{0.093}}$ \\
        
        & MMPolymer (ours) & {\textbf{0.431}$_{\pm\text{0.017}}$} & 
        {\textbf{0.496}$_{\pm\text{0.031}}$}  & {\textbf{0.286}$_{\pm\text{0.029}}$}  & {\textbf{0.390}$_{\pm\text{0.057}}$} & 
        {\textbf{16.814}$_{\pm\text{0.867}}$} & {\textbf{0.511}$_{\pm\text{0.035}}$} & 
        {\textbf{0.087}$_{\pm\text{0.010}}$} & {0.061}$_{\pm\text{0.016}}$ \\
    \midrule
    \multirow{9}[0]{*}{$R^2$ ($\uparrow$)} 
          & ChemBERTa~\cite{ahmad2022chemberta} & 0.880$_{\pm\text{0.023}}$ & 0.881$_{\pm\text{0.028}}$ & 0.888$_{\pm\text{0.035}}$ & 0.745$_{\pm\text{0.102}}$ & 0.365$_{\pm\text{0.098}}$ & 0.682$_{\pm\text{0.123}}$ & 0.643$_{\pm\text{0.076}}$ & 0.590$_{\pm\text{0.078}}$ \\

          & MolCLR~\cite{wang2022molecular} & 0.858$_{\pm\text{0.010}}$ & 0.882$_{\pm\text{0.027}}$ & 0.854$_{\pm\text{0.038}}$ & 0.689$_{\pm\text{0.037}}$ & 0.176$_{\pm\text{0.026}}$ & 0.683$_{\pm\text{0.020}}$ &
          0.764$_{\pm\text{0.037}}$ & 0.885$_{\pm\text{0.104}}$ \\

          & 3D Infomax~\cite{stark20223d} & 0.900$_{\pm\text{0.016}}$ & 0.898$_{\pm\text{0.018}}$ & 0.891$_{\pm\text{0.045}}$ & 0.766$_{\pm\text{0.086}}$ & 0.274$_{\pm\text{0.122}}$ & 0.690$_{\pm\text{0.063}}$ &
          0.797$_{\pm\text{0.086}}$ & 0.869$_{\pm\text{0.097}}$ \\

          & Uni-Mol~\cite{zhou2023unimol} & 0.901$_{\pm\text{0.013}}$ & 0.925$_{\pm\text{0.011}}$  & 0.901$_{\pm\text{0.027}}$ & 0.820$_{\pm\text{0.075}}$ & 0.454$_{\pm\text{0.079}}$ & 0.751$_{\pm\text{0.085}}$ & 0.828$_{\pm\text{0.072}}$ & 0.937$_{\pm\text{0.032}}$  \\
          
    
          & SML~\cite{zhang2023transferring} & 0.901$_{\pm\text{0.022}}$ & 0.920$_{\pm\text{0.029}}$ & {0.915}$_{\pm\text{0.015}}$ & 0.802$_{\pm\text{0.051}}$ & 0.340$_{\pm\text{0.125}}$ & 0.726$_{\pm\text{0.038}}$ & 0.812$_{\pm\text{0.058}}$ & {0.967}$_{\pm\text{0.015}}$ \\
    
          & PLM~\cite{zhang2023transferring} & 0.911$_{\pm\text{0.014}}$ & {0.925}$_{\pm\text{0.021}}$ & 0.910$_{\pm\text{0.019}}$ & 0.791$_{\pm\text{0.049}}$ & 0.330$_{\pm\text{0.105}}$ & 0.726$_{\pm\text{0.058}}$ & 0.817$_{\pm\text{0.056}}$ &{\textbf{0.980}}$_{\pm\text{0.008}}$ \\
                    
          & polyBERT~\cite{kuenneth2023polybert} & 0.875$_{\pm\text{0.006}}$ & 0.844$_{\pm\text{0.034}}$ & 0.880$_{\pm\text{0.035}}$ & 0.705$_{\pm\text{0.085}}$ & 0.384$_{\pm\text{0.066}}$ & 0.681$_{\pm\text{0.058}}$ & 0.769$_{\pm\text{0.034}}$ & 0.672$_{\pm\text{0.119}}$ \\

          & Transpolymer~\cite{xu2023transpolymer} &
         {0.916}$_{\pm\text{0.002}}$ & 0.911$_{\pm\text{0.008}}$ & 0.902$_{\pm\text{0.036}}$ & {0.830}$_{\pm\text{0.059}}$ & {0.430}$_{\pm\text{0.058}}$ & {0.744}$_{\pm\text{0.075}}$ & {0.826}$_{\pm\text{0.071}}$ & 0.800$_{\pm\text{0.172}}$ \\
          
          & MMPolymer (ours) & 
          {\textbf{0.924}}$_{\pm\text{0.006}}$ & {\textbf{0.934}}$_{\pm\text{0.008}}$ & {\textbf{0.925}}$_{\pm\text{0.025}}$ & {\textbf{0.836}}$_{\pm\text{0.053}}$ & 
          {\textbf{0.488}}$_{\pm\text{0.072}}$ & {\textbf{0.779}}$_{\pm\text{0.052}}$ & {\textbf{0.864}}$_{\pm\text{0.036}}$ & 0.961$_{\pm\text{0.018}}$ \\
    \bottomrule
    \end{tabular}
  \label{tab:main_exp}
\end{table*}%

\begin{figure*}[htp]
  \centering
  \subfigure[Egc]{
    \includegraphics[height=3.1cm]{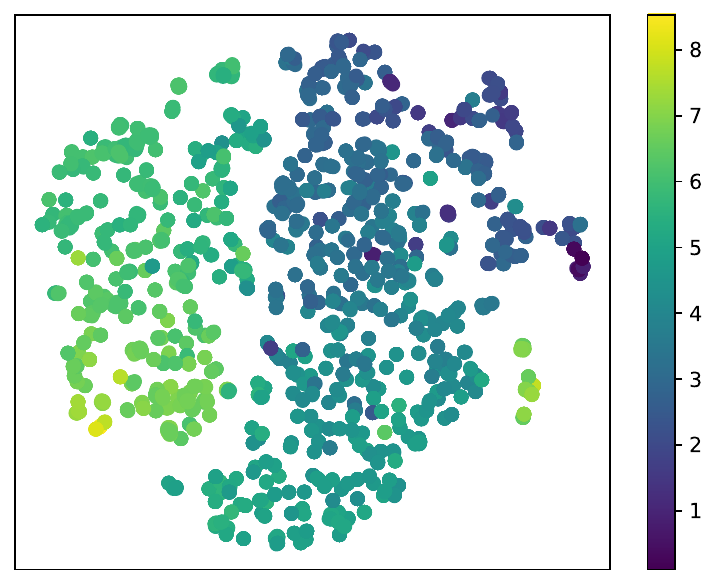}
  }
  \hspace*{0.05cm}
  \subfigure[Egb]{
    \includegraphics[height=3.1cm]{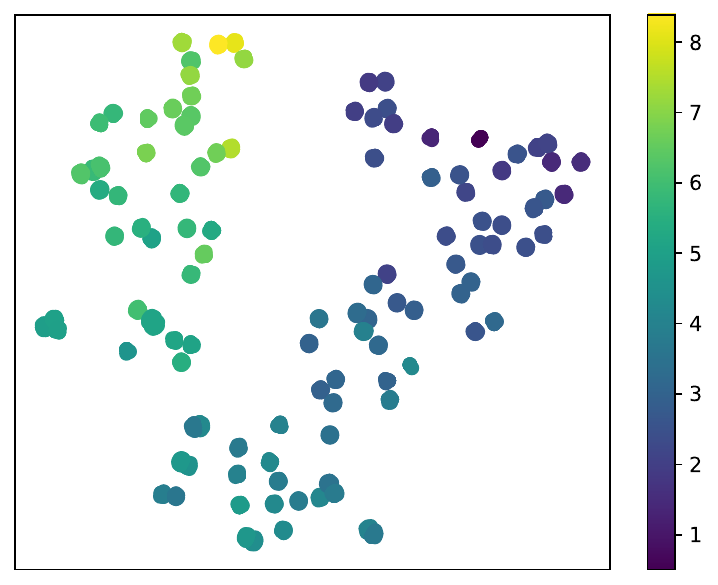}
  }
  \subfigure[Eea]{
    \includegraphics[height=3.1cm]{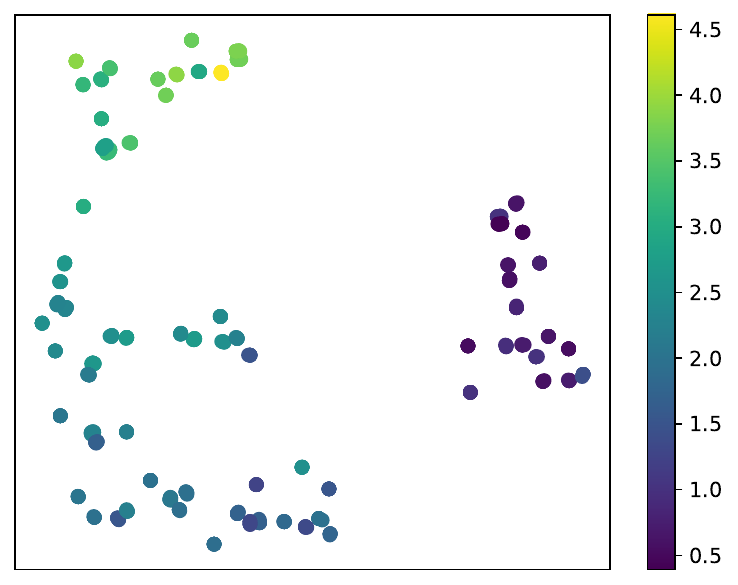}
  }
  \subfigure[Ei]{
    \includegraphics[height=3.1cm]{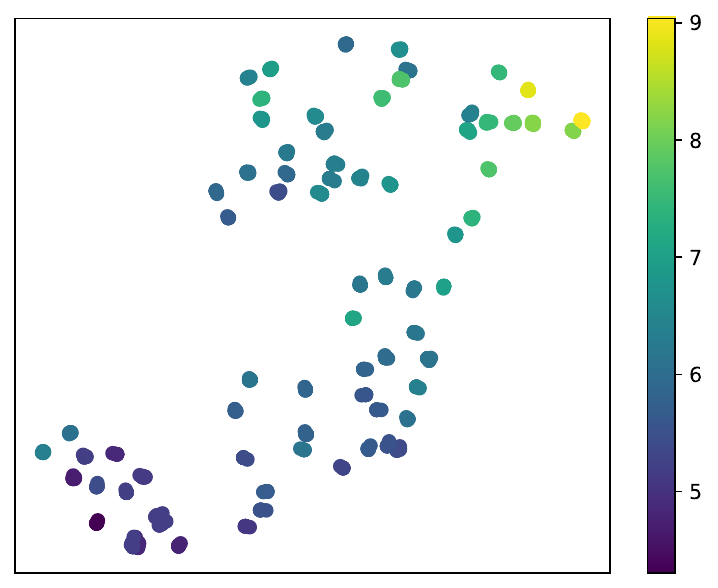}
  }
  \hspace*{0.2cm}
  \subfigure[Xc]{
    \includegraphics[height=3.1cm]{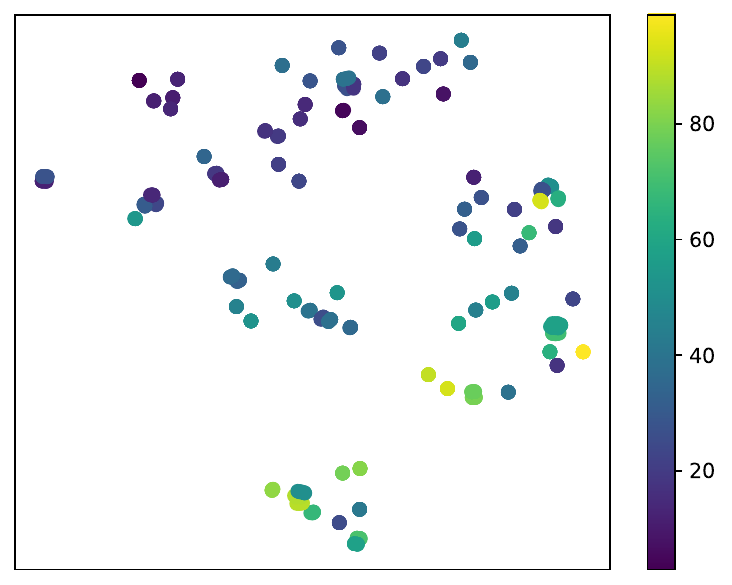}
  }
  \subfigure[EPS]{
    \includegraphics[height=3.1cm]{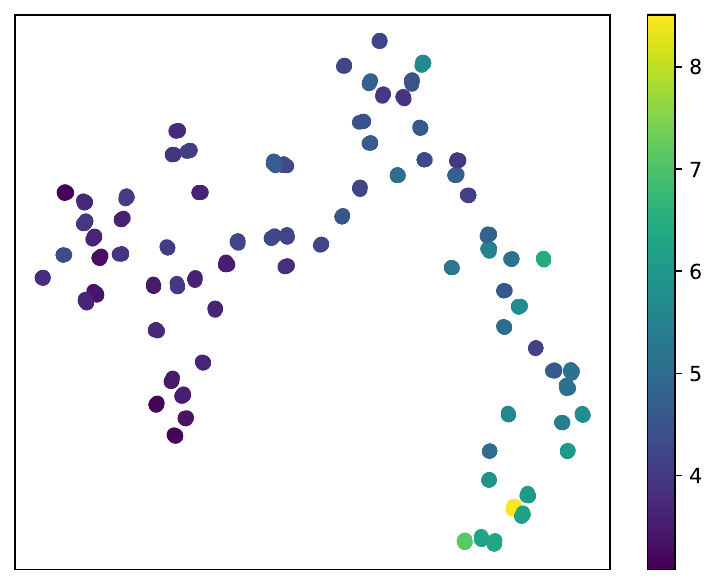}
  }
  \subfigure[Nc]{
    \includegraphics[height=3.1cm]{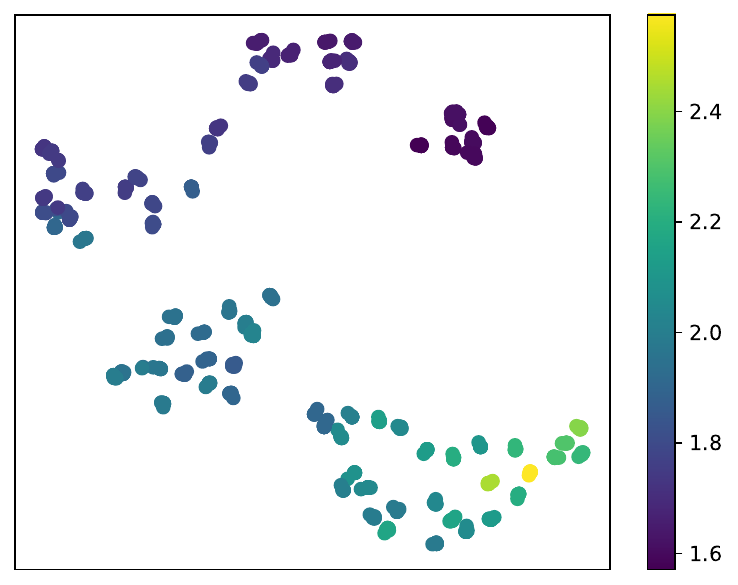}
  }
  \subfigure[Eat]{
    \includegraphics[height=3.1cm]{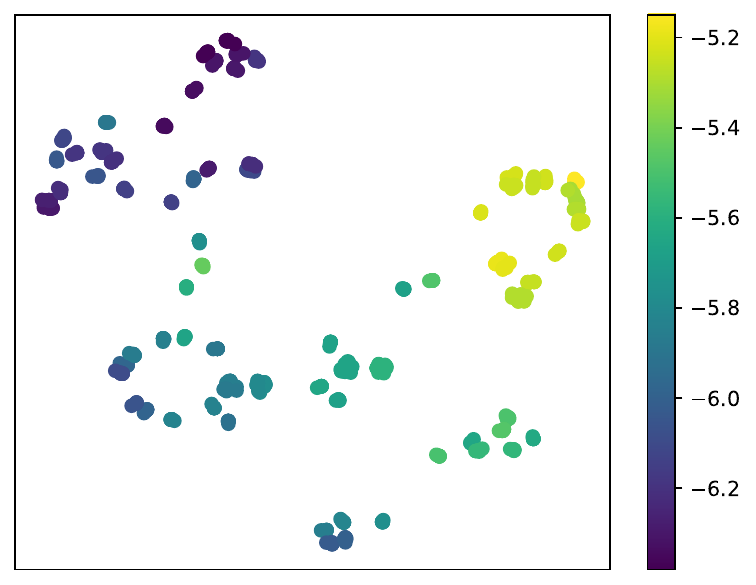}
  }
  \caption{t-SNE visualization of MMPolymer on eight polymer property datasets, where the color for each data point is determined by the corresponding ground truth (i.e., property label).} 
  \label{fig:tsne}
\end{figure*}

\begin{table}[htbp]
  \centering
  \caption{The performance improvement of MMPolymer-1D, which only utilizes polymer 1D sequential information during fine-tuning, and MMPolymer-3D, which only utilizes 3D structural information during fine-tuning, compared with the best baseline on each polymer property dataset.
  }
    \begin{tabular}{c|cc|cc}
    \toprule
    \multirow{2}[4]{*}{Dataset} & \multicolumn{2}{c|}{MMPolymer-1D} & \multicolumn{2}{c}{MMPolymer-3D} \\
\cmidrule{2-5}   & $\Delta$RMSE ($\downarrow$) & $\Delta R^2$ ($\uparrow$)  & $\Delta$RMSE ($\downarrow$) & $\Delta R^2$ ($\uparrow$) \\
    \midrule
    Egc   & \textbf{-0.008} & \textbf{+0.003} & \textbf{-0.004} & \textbf{+0.001} \\
    Egb   & \textbf{-0.015} & \textbf{+0.005} & \textbf{-0.025} & \textbf{+0.007} \\
    Eea   & \textbf{-0.009} & \textbf{+0.002} & \textbf{-0.027} & \textbf{+0.010} \\
    Ei    & +0.016 & -0.011 & \textbf{-0.003} & \textbf{+0.004} \\
    Xc    & \textbf{-0.219} & \textbf{+0.010} & +0.449 & -0.028 \\
    EPS   & \textbf{-0.011} & \textbf{+0.016} & \textbf{-0.025} & \textbf{+0.028} \\
    Nc    & \textbf{-0.005} & \textbf{+0.024} & \textbf{-0.006} & \textbf{+0.025} \\
    Eat   & +0.028 & -0.046 & +0.011 & -0.019 \\
    \bottomrule
    \end{tabular}%
  \label{tab:single_modal}%
\end{table}%

\subsubsection{Baselines}
To demonstrate the effectiveness of our method, several state-of-the-art methods have been compared, including \textbf{SML}~\cite{zhang2023transferring}, \textbf{PLM}~\cite{zhang2023transferring}, \textbf{polyBERT}~\cite{kuenneth2023polybert} and \textbf{Transpolymer}~\cite{xu2023transpolymer}, which are all pretraining-based methods for polymer property prediction.
Methods lacking pretraining, like GP~\cite{kuenneth2021polymer}, have already been surpassed by current pretraining-based methods~\cite{xu2023transpolymer}, hence we no longer include them in our comparisons. 
Besides, we also compare our method with four representative molecular pretraining methods, 
including \textbf{ChemBERTa}~\cite{ahmad2022chemberta}, ~\textbf{MolCLR}~\cite{wang2022molecular}, ~\textbf{3D Infomax}~\cite{stark20223d} and ~\textbf{Uni-Mol}~\cite{zhou2023unimol}, to reveal the differences between small molecules and polymers. 
The baselines are implemented based on their default settings in the corresponding references.

\subsubsection{Evaluation Metircs}
Since polymer property prediction tasks on the fine-tuning datasets are all regression tasks, we also choose root mean squared error (RMSE) and R-squared ($R^2$) as evaluation metrics in line with previous works~\cite{xu2023transpolymer, kuenneth2023polybert, zhang2023transferring}, thus guaranteeing a comprehensive and fair assessment of the predictive performance.

\subsubsection{Model architecture and hyperparameter settings}
Our 1D representation network consists of 6 encoding layers, with 12 attention heads each, whereas our 3D representation network comprises 15 encoding layers, with 64 attention heads each.
Besides, our model is pretrained on the eight Tesla V100 GPUs using a batch size of 16 while the adam optimizer~\cite{kingma2014adam} is utilized with a learning rate of 1e-4, betas of \((0.9, 0.99)\), and eps of 1e-6.

\begin{table*}[h]
\caption{The performance of MMPolymer under different data processing strategies, where the "Star Keep" strategy refers to directly using the original P-SMILES string, the "Star Remove" strategy refers to removing "*" symbols in the original P-SMILES string and the "Star Substitution" strategy refers to replacing the "*" symbol in the original P-SMILES string with the neighboring atom symbol of
another "*" symbol.} 
  \centering
  \setlength{\tabcolsep}{2.5pt}
    \centering\begin{tabular}{c|c|cccccccc}
    \toprule
    Metric & Strategy & Egc   & Egb   & Eea   & Ei    & Xc    & EPS   & Nc    & Eat \\
    \midrule
    \multirow{3}[0]{*}{RMSE ($\downarrow$)} 
        & Star Keep & 0.433$_{\pm\text{0.015}}$ & 0.506$_{\pm\text{0.055}}$ & 0.310$_{\pm\text{0.022}}$ & {0.392$_{\pm\text{0.047}}$} & 16.836$_{\pm\text{1.272}}$ & 0.514$_{\pm\text{0.047}}$ &
        0.089$_{\pm\text{0.011}}$ & 0.077$_{\pm\text{0.042}}$ \\

        & Star Remove & 0.478$_{\pm\text{0.023}}$ & 0.527$_{\pm\text{0.012}}$  & 0.299$_{\pm\text{0.016}}$ & 0.403$_{\pm\text{0.041}}$ & 16.962$_{\pm\text{0.803}}$ & 0.514$_{\pm\text{0.032}}$ & 0.091$_{\pm\text{0.011}}$ & 0.083$_{\pm\text{0.040}}$  \\

        & Star Substitution (Ours)  & {\textbf{0.431}$_{\pm\text{0.017}}$} & 
        {\textbf{0.496}$_{\pm\text{0.031}}$}  & {\textbf{0.286}$_{\pm\text{0.029}}$}  & {\textbf{0.390}$_{\pm\text{0.057}}$} & 
        {\textbf{16.814}$_{\pm\text{0.867}}$} & {\textbf{0.511}$_{\pm\text{0.035}}$} & 
        {\textbf{0.087}$_{\pm\text{0.010}}$} & 
        {\textbf{0.061}$_{\pm\text{0.016}}$} \\
        
    \midrule
    \multirow{3}[0]{*}{$R^2$ ($\uparrow$)} 
          & Star Keep & 0.921$_{\pm\text{0.006}}$ & 0.927$_{\pm\text{0.008}}$ & 0.914$_{\pm\text{0.022}}$ & 0.833$_{\pm\text{0.049}}$ & 0.483$_{\pm\text{0.102}}$ & 0.778$_{\pm\text{0.055}}$ &
          0.855$_{\pm\text{0.038}}$ & 0.945$_{\pm\text{0.041}}$ \\

          & Star Remove & 0.905$_{\pm\text{0.009}}$ & 0.925$_{\pm\text{0.013}}$ & 0.921$_{\pm\text{0.015}}$ & 0.828$_{\pm\text{0.043}}$ & 0.471$_{\pm\text{0.093}}$ & 0.776$_{\pm\text{0.050}}$ & 0.845$_{\pm\text{0.053}}$ & 0.938$_{\pm\text{0.040}}$ \\
          
          & Star Substitution (Ours) & 
          {\textbf{0.924}}$_{\pm\text{0.006}}$ & {\textbf{0.934}}$_{\pm\text{0.008}}$ & {\textbf{0.925}}$_{\pm\text{0.025}}$ & {\textbf{0.836}}$_{\pm\text{0.053}}$ & 
          {\textbf{0.488}}$_{\pm\text{0.072}}$ & {\textbf{0.779}}$_{\pm\text{0.052}}$ & {\textbf{0.864}}$_{\pm\text{0.036}}$ & {\textbf{0.961}}$_{\pm\text{0.018}}$ \\
    \bottomrule
    \end{tabular}
  \label{tab:abla_exp1}
\end{table*}%

\begin{table*}[h]
  \centering
  \caption{The performance of MMPolymer under different pretraining settings, where "1D pre" refers to the masking prediction task on the 1D representation network, "3D pre" refers to the coordinate denoising task on the 3D representation network, and "Contrast" refers to the cross-modal alignment task.} 
  \setlength{\tabcolsep}{2pt}
    \begin{tabular}{c|ccc|cccccccc}
    \toprule
    Metric &
    1D pre & 
    3D pre & 
    Contrast & Egc   & Egb   & Eea   & Ei    & Xc    & EPS   & Nc    & Eat \\
    \midrule
        \multirow{8}[0]{*}{RMSE ($\downarrow$)} 
    & $\times$ & $\times$ & $\times$ & 0.596$_{\pm\text{0.022}}$ & 0.575$_{\pm\text{0.031}}$ & 0.343$_{\pm\text{0.029}}$ & 0.432$_{\pm\text{0.062}}$ & 20.152$_{\pm\text{1.624}}$ & 0.569$_{\pm\text{0.053}}$ & 0.102$_{\pm\text{0.010}}$ & 0.122$_{\pm\text{0.051}}$ \\
    
    & $\checkmark$ & $\times$ & $\times$ & {0.438}$_{\pm\text{0.010}}$ & 0.510$_{\pm\text{0.034}}$ & 0.311$_{\pm\text{0.028}}$ & {0.394}$_{\pm\text{0.058}}$ & {{16.818}}$_{\pm\text{0.779}}$ & 0.535$_{\pm\text{0.052}}$ & {0.091}$_{\pm\text{0.008}}$ & {0.112}$_{\pm\text{0.085}}$ \\

     & $\times$ & $\checkmark$ & $\times$ &  0.492$_{\pm\text{0.026}}$ & 0.542$_{\pm\text{0.042}}$ & 0.337$_{\pm\text{0.023}}$ & 0.424$_{\pm\text{0.080}}$ & 18.542$_{\pm\text{0.849}}$ & 0.532$_{\pm\text{0.049}}$ & 0.096$_{\pm\text{0.015}}$ & 0.086$_{\pm\text{0.030}}$ \\

     & $\times$ & $\times$ & $\checkmark$ & 0.618$_{\pm\text{0.013}}$ & 0.716$_{\pm\text{0.058}}$ & 0.392$_{\pm\text{0.041}}$ & 0.456$_{\pm\text{0.079}}$ & 19.624$_{\pm\text{2.003}}$ & 0.632$_{\pm\text{0.031}}$ & 0.119$_{\pm\text{0.015}}$ & 0.162$_{\pm\text{0.094}}$ \\

     & $\checkmark$ & $\times$ & $\checkmark$ & 0.459$_{\pm\text{0.010}}$ & 0.519$_{\pm\text{0.044}}$ & 0.298$_{\pm\text{0.032}}$ & 0.417$_{\pm\text{0.048}}$ & 17.033$_{\pm\text{0.482}}$ & 0.545$_{\pm\text{0.062}}$ & 0.092$_{\pm\text{0.012}}$ & 0.124$_{\pm\text{0.073}}$ \\

     & $\times$ & $\checkmark$ & $\checkmark$ & 0.593$_{\pm\text{0.018}}$ & 0.574$_{\pm\text{0.037}}$ & 0.347$_{\pm\text{0.032}}$ & 0.421$_{\pm\text{0.053}}$ & 17.816$_{\pm\text{1.417}}$ & 0.560$_{\pm\text{0.075}}$ & 0.104$_{\pm\text{0.011}}$ & 0.140$_{\pm\text{0.056}}$ \\
     
     & $\checkmark$ & $\checkmark$ & $\times$ & {0.440}$_{\pm\text{0.017}}$ & {{0.502}}$_{\pm\text{0.051}}$ & {0.308}$_{\pm\text{0.030}}$ & 0.400$_{\pm\text{0.062}}$ & 17.177$_{\pm\text{0.554}}$ & {0.532}$_{\pm\text{0.055}}$ & 0.092$_{\pm\text{0.011}}$ & 0.115$_{\pm\text{0.084}}$ \\

     & $\checkmark$ & $\checkmark$ & $\checkmark$ &  
    {\textbf{0.431}}$_{\pm\text{0.017}}$  & \textbf{0.496}$_{\pm\text{0.031}}$ & {\textbf{0.286}$_{\pm\text{0.029}}$} & {\textbf{0.390}}$_{\pm\text{0.057}}$ & \textbf{16.814}$_{\pm\text{0.867}}$ & \textbf{0.511}$_{\pm\text{0.035}}$ & {\textbf{0.087}}$_{\pm\text{0.010}}$ & {\textbf{0.061}}$_{\pm\text{0.016}}$ \\

    \midrule
    \multirow{8}[0]{*}{$R^2$ ($\uparrow$)}
     & $\times$ & $\times$ & $\times$ &  0.854$_{\pm\text{0.010}}$ & 0.910$_{\pm\text{0.019}}$ & 0.895$_{\pm\text{0.025}}$ & 0.799$_{\pm\text{0.064}}$ & 0.269$_{\pm\text{0.090}}$ & 0.725$_{\pm\text{0.079}}$ & 0.811$_{\pm\text{0.040}}$ & 0.873$_{\pm\text{0.082}}$ \\

     & $\checkmark$ & $\times$ & $\times$ &  {0.921}$_{\pm\text{0.005}}$ & 0.931$_{\pm\text{0.003}}$ & 0.914$_{\pm\text{0.020}}$ & {0.833}$_{\pm\text{0.054}}$ & {\textbf {0.488}}$_{\pm\text{0.061}}$ & 0.757$_{\pm\text{0.065}}$ & 0.847$_{\pm\text{0.050}}$ & {0.878}$_{\pm\text{0.137}}$ \\

     & $\times$ & $\checkmark$ & $\times$ &  0.900$_{\pm\text{0.012}}$ & 0.921$_{\pm\text{0.013}}$ & 0.897$_{\pm\text{0.029}}$ & 0.803$_{\pm\text{0.079}}$ & 0.378$_{\pm\text{0.077}}$ & 0.758$_{\pm\text{0.069}}$ & 0.824$_{\pm\text{0.080}}$ & 0.937$_{\pm\text{0.026}}$ \\

     & $\times$ & $\times$ & $\checkmark$ &  0.843$_{\pm\text{0.002}}$ & 0.862$_{\pm\text{0.022}}$ & 0.862$_{\pm\text{0.037}}$ & 0.773$_{\pm\text{0.090}}$ & 0.303$_{\pm\text{0.125}}$ & 0.665$_{\pm\text{0.060}}$ & 0.744$_{\pm\text{0.058}}$ & 0.753$_{\pm\text{0.158}}$ \\

     & $\checkmark$ & $\times$ & $\checkmark$ & 0.913$_{\pm\text{0.004}}$ & 0.928$_{\pm\text{0.006}}$ & 0.919$_{\pm\text{0.024}}$ & 0.816$_{\pm\text{0.047}}$ & 0.474$_{\pm\text{0.067}}$ & 0.741$_{\pm\text{0.094}}$ & 0.837$_{\pm\text{0.066}}$ & 0.865$_{\pm\text{0.107}}$ \\

     & $\times$ & $\checkmark$ & $\checkmark$ & 0.855$_{\pm\text{0.007}}$ & 0.910$_{\pm\text{0.021}}$ & 0.892$_{\pm\text{0.026}}$ & 0.810$_{\pm\text{0.056}}$ & 0.422$_{\pm\text{0.105}}$ & 0.730$_{\pm\text{0.093}}$ & 0.803$_{\pm\text{0.042}}$ & 0.827$_{\pm\text{0.087}}$ \\

     & $\checkmark$ & $\checkmark$ & $\times$ & {0.921}$_{\pm\text{0.005}}$ & {{0.933}}$_{\pm\text{0.009}}$ & {0.915}$_{\pm\text{0.020}}$ & 0.827$_{\pm\text{0.060}}$ & {0.466}$_{\pm\text{0.060}}$ &  {0.758}$_{\pm\text{0.072}}$ & {0.847}$_{\pm\text{0.034}}$ & 0.871$_{\pm\text{0.135}}$ \\
    
     & $\checkmark$ & $\checkmark$ & $\checkmark$ &  {\textbf{0.924}}$_{\pm\text{0.006}}$ & \textbf{0.934}$_{\pm\text{0.008}}$ & {\textbf{0.925}}$_{\pm\text{0.025}}$ & {\textbf{0.836}}$_{\pm\text{0.053}}$ & \textbf{0.488}$_{\pm\text{0.072}} $ & {\textbf{0.779}}$_{\pm\text{0.052}}$ & {\textbf{0.864}}$_{\pm\text{0.036}}$ & {\textbf{0.961}}$_{\pm\text{0.018}}$ \\

    \bottomrule
    \end{tabular}
  \label{tab:abla_exp2}
\end{table*}%

\subsection{Comparisons}
\subsubsection{Demonstration of effectiveness} 
Table~\ref{tab:main_exp} presents the performance of different methods on the eight polymer property datasets, which are evaluated by RMSE and $R^2$.
In general, our proposed method, MMPolymer, stands out by achieving state-of-the-art performance on seven datasets and comparable performance on the remaining Eat dataset, underscoring its effectiveness on polymer property prediction tasks.
Notably, several previous works, such as Transpolymer~\cite{xu2023transpolymer}, have excluded the Eat dataset from their original paper because it fails to capture the complete polymer structure\footnote{Eat is a localized property, which depends on the local atomic environment.} and therefore leads to lower reliability compared to other property datasets.
Besides, even if considering the Eat dataset, the superiority of our MMPolymer is still obvious.

Moreover, comparing the four representative molecular pretraining methods reveals some insightful trends.
On the one hand, the 3D molecular pretraining method Uni-Mol consistently outperforms the 1D molecular pretraining method ChemBERTa across all eight polymer property datasets, emphasizing the crucial role of 3D structural information in enhancing predictive performance.
On the other hand, all molecular pretraining methods exhibit inferior performance when compared with our MMPolymer.
These observations indicate the inherent limitations of directly applying molecular pretraining methods to polymer-specific tasks, emphasizing the significance of tailored methods like our MMPolymer for achieving accurate polymer property prediction.

As shown in Figure~\ref{fig:tsne}, we conduct t-SNE visualization~\cite{van2008visualizing} based on the polymer representations learned by our MMPolymer and corresponding ground truth for each polymer property dataset. 
The t-SNE visualization illustrates that the chemical space defined by the representations accurately captures the differences in the ground truth for each data point, as reflected by the color gradations. 
These colors are determined by the corresponding ground truth, and the spread of colors across the datasets indicates a diverse range of values. 
This variation confirms that the representations learned by our MMPolymer are sensitive to the underlying ground truth differences among data points. 
The distinct clusters, especially in datasets like Egb, Eea, and Eat, where the color intensity marks out separate regions, serve as a testament to the fidelity with which the representations map out the actual property landscape of the dataset. 
The linear color variation further emphasizes that the representations are not only capturing distinct property values but are also able to reflect a continuum in the property and chemical space, showing the superiority of our MMPolymer.

\subsubsection{Adaptability for various downstream settings}
We also explore the predictive capacity of our MMPolymer when only utilizing single modality information (either polymer 1D sequential information or 3D structural information) during fine-tuning.
As shown in Table~\ref{tab:single_modal}, compared to the best baseline on each polymer property dataset, both MMPolymer-1D (i.e., only utilize polymer 1D sequential information during fine-tuning) and MMPolymer-3D (i.e., only utilize polymer 3D structural information during fine-tuning) consistently achieve lower RMSE and higher $R^2$ on most datasets.
This finding highlights the adaptability of our MMPolymer, demonstrating that it has acquired adequate general knowledge through our multimodal multitask pretraining paradigm, enabling MMPolymer to flexibly choose modality information based on the characteristics of the property being predicted and the specific requirements of the corresponding fine-tuning task.

\subsection{Ablation Studies}
\subsubsection{Star Substitution Strategy}\label{section:Abla_star} 

As mentioned in Sec.~\ref{section:3D_net}, considering the regularity of polymer 3D structure and scarcity of polymer 3D data, we apply the 3D conformation of its repeating unit (represented by the P-SMILES string) to approximate the whole polymer 3D conformation.
Particularly, before generating this 3D conformation, we first convert the original P-SMILES string through our "Star Substitution" strategy, where the "*" symbol in the original P-SMILES string is replaced with the neighboring atom symbol of another "*" symbol, to reflect the structural features of the whole polymer 3D conformation as more as possible.
Here, comprehensive ablation studies have been conducted to evaluate the effectiveness of our "Star Substitution" strategy.

Table~\ref{tab:abla_exp1} displays the performance of MMPolymer under different data processing strategies, including the default "Star keep" strategy (i.e., directly uses the original P-SMILES string), the "Star Remove" strategy (i.e., remove "*" symbols in the original P-SMILES string), and our "Star Substitution" strategy (i.e.,  replace the "*" symbol in the original P-SMILES string with the neighboring atom symbol of another "*" symbol).
In general, compared with the default "Star keep" strategy and the "Star Remove" strategy, the performance of MMPolymer can be consistently improved on all eight polymer property datasets by using our "Star Substitution" strategy.
This indicates that the 3D conformation of the repeating unit can capture more structural features of the whole polymer 3D conformation through our "Star Substitution" strategy, thus emphasizing the effectiveness of our "Star Substitution" strategy.

In particular, the performance of MMPolymer degrades when using the "Star Remove" strategy.
We attribute this phenomenon to the lack of polymerization site information.
As mentioned in Sec.~\ref{section:1D_net}, the P-SMILES string is formed by combining the SMILES string of the corresponding monomer with two "*" symbols, where "*" symbols are used to indicate the polymerization sites.
Therefore, removing "*" symbols means using the monomer's 3D conformation rather than the repeating unit's 3D conformation.
In this case, it's difficult to capture the structural features of the whole polymer 3D conformation, thus leading to worse performance.
This phenomenon also further indicates the specificity of polymers and the need for tailored methods rather than directly transferring methods from other domains (e.g., protein domain or molecule domain).

\subsubsection{Multimodal Multitask Pretraining Paradigm} 
As mentioned in Sec.~\ref{section:learning_paradigm}, MMPolymer is pretrained through our multimodal multitask paradigm, including the masked prediction task for pretraining the 1D representation network, the coordinate denoising task for pretraining the 3D representation network, and the cross-modal alignment task for aligning the learned multimodal representation.
Here, to evaluate the effectiveness of our multimodal multitask pretraining paradigm, assessing the specific impact of 1D pretraining (i.e., the masked prediction task), 3D pretraining (i.e., the coordinate denoising task), and contrastive learning (i.e., the cross-modal alignment task) on the model performance, comprehensive ablation studies have been conducted.

Table~\ref{tab:abla_exp2} displays the performance of MMPolymer under different pretraining settings, thus providing valuable insights into our proposed multimodal multitask pretraining paradigm.
In general, when all three pretraining tasks (i.e., masked prediction task, coordinate denoising task, and cross-modal alignment task) are combined, MMPolymer achieves its optimal performance.
This underscores the synergistic effect of our multimodal multitask pretraining paradigm, resulting in a robust and powerful model capable of capturing diverse aspects of polymer data.

Specifically, we observe that incorporating 3D structural information greatly enhances the performance of MMPolymer, especially when combining both 1D sequential and 3D structural information, showcasing the critical role of 3D structural information in polymer property prediction.
Moreover, the utilization of contrastive learning further enhances the performance of MMPolymer. 
This suggests that aligning learned representations across different modalities contributes to better multimodal integration, thus improving the predictive performance of polymer properties.
Notably, we aim to leverage complementary 1D and 3D information for polymer property prediction, rather than matching the two modalities. 
In this case, purely applying the contrastive learning task is insufficient and leads to unsatisfactory performance, as shown in Table~\ref{tab:abla_exp2}.

Besides, we can observe that the performance of MMPolymer with only 1D pretraining surpasses that with 3D pretraining on most datasets.
We attribute this phenomenon to the fact that we approximate the whole polymer 3D conformation with the 3D conformation of its corresponding repeating unit.
Although through our "Star Substitution" strategy, as analyzed in Sec.~\ref{section:Abla_star}, the 3D conformation of the repeating unit can reflect the structural features of the whole polymer 3D conformation to some extent, it is still not accurate enough to describe all natures.
This observation highlights the current issue of insufficient high-precision polymer 3D structure data and indicates that pure 3D pretraining methods still face significant challenges in the polymer domain, further emphasizing the benefits of our multimodal multitask pretraining paradigm.

\section{Conclusions}
In this work, we present MMPolymer, a multimodal multitask pretraining framework, to achieve accurate polymer property prediction\footnote{Code is available at \url{https://github.com/FanmengWang/MMPolymer}}. 
By effectively combining polymer 1D sequential and 3D structural information through our multimodal multitask pretraining paradigm, MMPolymer can fully capture diverse aspects of polymer data, creating a robust and promising model for downstream polymer property prediction tasks.
Besides, through our "Star Substitution" strategy, the 3D structural information can be extracted effectively, overcoming the scarcity of polymer 3D data to some extent.
The extensive experiments consistently demonstrate that MMPolymer achieves state-of-the-art performance on various polymer property prediction tasks, significantly outperforming existing polymer property prediction methods.
Even if only single-modal information (either polymer 1D sequential or 3D structural information) is utilized in the fine-tuning phase, the pretrained MMPolymer can still surpass existing polymer property prediction methods, showcasing its exceptional capability in polymer feature extraction and utilization.
Moreover, comprehensive ablation studies are also conducted to confirm the rationality and effectiveness of our proposed method.
For future work, we will continue to explore the intrinsic relationship between polymer structures and their properties, achieving better modeling of polymer structures to benefit its many downstream applications.

\begin{acks}
This work was supported in part by the National Natural Science Foundation of China (62106271, 92270110), the Fundamental Research Funds for the Central Universities, and the Research Funds of Renmin University of China. 
Dr. Hongteng Xu thanks the support from the Beijing Key Laboratory of Big Data Management and Analysis Methods, the Intelligent Social Governance Platform, Major Innovation \& Planning Interdisciplinary Platform for the ``Double-First Class'' Initiative. 
\end{acks}

\bibliographystyle{ACM-Reference-Format}
\balance
\bibliography{sample-base}

\end{document}